\documentclass[10pt,twocolumn,letterpaper]{article}

\usepackage{iccv}
\usepackage{times}
\usepackage{epsfig}
\usepackage{graphicx}
\usepackage{amsmath}
\usepackage{amssymb}
\usepackage{multirow}
\usepackage{caption}
\usepackage{subcaption}
\usepackage{booktabs} 
\usepackage[boxed,ruled]{algorithm2e}
\usepackage{colortbl}
\usepackage{tikz}
\usepackage{lipsum}
\usepackage{soul}

\newcommand{\cut}[1]{}

\definecolor{myGreen}{HTML}{3FAA59}

\usepackage[pagebackref=true,breaklinks=true,letterpaper=true,colorlinks,bookmarks=false]{hyperref}

\iccvfinalcopy 

\def\iccvPaperID{1858} 

\ificcvfinal\fi

\DeclareMathOperator*{\argmin}{arg\,min}
\DeclareMathOperator*{\argmax}{arg\,max}

\begin{document}

\title{Neural Fine-Tuning Search for Few-Shot Learning}

\newcommand{\longName}{Neural Fine-Tuning Search}
\newcommand{\shortName}{NFTS}

\author{
Panagiotis Eustratiadis\textsuperscript{1}
\and
Łukasz Dudziak\textsuperscript{2}
\and
Da Li\textsuperscript{1,2}
\and
Timothy Hospedales\textsuperscript{1,2}
\and
\textsuperscript{1}University of Edinburgh\\
\and
\textsuperscript{2}Samsung AI Center, Cambridge\\
}

\maketitle
\ificcvfinal\thispagestyle{empty}\fi

\begin{abstract}
In few-shot recognition, a classifier that has been trained on one set of classes is required to rapidly adapt and generalize to a disjoint, novel set of classes. To that end, recent studies have shown the efficacy of fine-tuning with carefully crafted adaptation architectures. However this raises the question of: How can one design the optimal adaptation strategy? In this paper, we study this question through the lens of neural architecture search (NAS). Given a pre-trained neural network, our algorithm discovers the optimal arrangement of adapters, which layers to keep frozen and which to fine-tune. We demonstrate the generality of our NAS method by applying it to both residual networks and vision transformers and report state-of-the-art performance on Meta-Dataset and Meta-Album.
\end{abstract}

\begin{figure}[t]
  \centering
  \includegraphics[width=\linewidth]{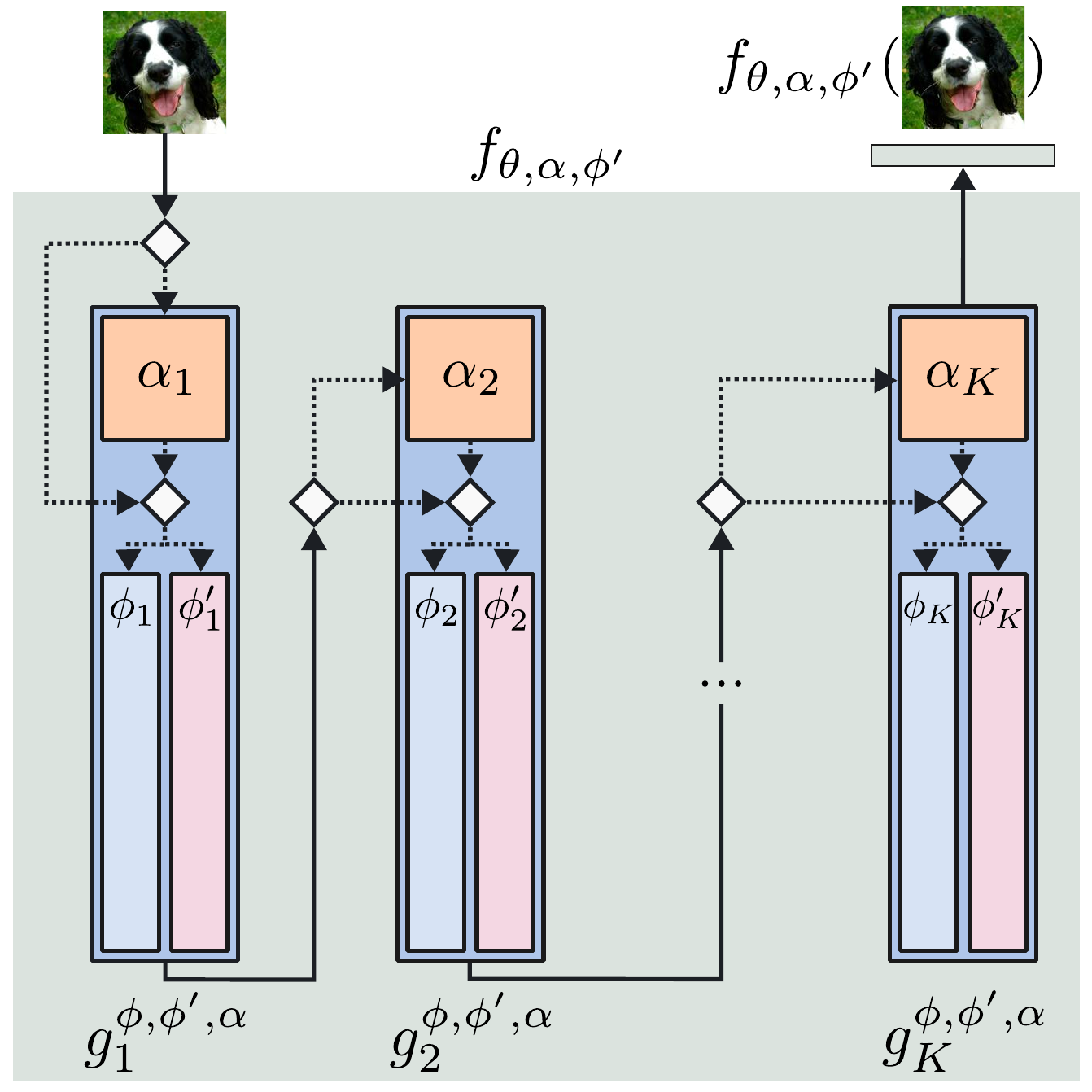}
  \caption{Our proposed supernet architecture for few-shot adaptation. The supernet contains all combinations of pre-trained, fine-tuned and adapter parameters. $f$ denotes the feature extractor, which is composed of many layers, $g$, which are the minimal unit for adaptation in our search space. The dotted lines represent possible paths that can be sampled during SPOS training. Every adaptable layer $g_i^{\phi,\phi',\alpha}$ has its own pre-trained parameters ($\phi_i \subset \theta$), fine-tuned parameters ($\phi'_i$), and adapter parameters ($\alpha_i$).}
  \label{fig:supernet-architecture}
  \vspace{-1em}
\end{figure}

\section{Introduction}
\label{sec:introduction}

Few-shot recognition~\cite{cogsci11_one_shot,cvpr00_one_shot,acm20_fsl_survey} aims to learn novel concepts from few examples, often by rapid adaptation of a model trained on a disjoint set of labels. Many solutions adopt a meta-learning perspective~\cite{icml17_maml,cvpr19_convex,iclr17opt,iclr19_embedding,neurips17_protonet}, or train a powerful feature extractor on the source classes~\cite{tian2020rethinking,wang2019simpleshot} -- both of which assume that the training and testing classes are drawn from the same underlying distribution e.g., handwritten characters~\cite{science15_omniglot}, or ImageNet categories \cite{neurips16_matching}. Later work considers a more realistic and challenging problem variant where a classifier should perform few-shot adaptation not only across visual categories, but also across diverse visual domains~\cite{iclr20_meta_dataset,neurips22_meta_album}. In this cross-domain problem variant, customising the feature extractor to the novel domains is important, and several studies address this through dynamic feature extractors~\cite{cvpr20_simple_cnaps,neurips19_cnaps} or ensembles of features~\cite{cvpr20_relevant,iccv21_url,iclr21_urt}. Another group of studies employ simple yet effective fine-tuning strategies for adaptation~\cite{dhillon2020baselineFSL,cvpr22_pmf,cvpr22_tsa,tmlr22_ett} that are predominantly heuristically motivated. Thus, an important question that arises from previous work is: How can one design the \textit{optimal} adaptation strategy? In this paper, we take a step towards answering this question.

Fine-tuning approaches to few-shot adaptation must manage a trade-off between adapting a large or small number of parameters. The former allows for better adaptation, but risks overfitting on a few-shot training set. The latter reduces the risk of overfitting, but limits the capacity for adaptation to novel categories and domains. The recent PMF~\cite{cvpr22_pmf} manages this trade-off through careful tuning of learning rates while fine-tuning the entire feature extractor. TSA~\cite{cvpr22_tsa} and ETT~\cite{tmlr22_ett} manage it by freezing the feature extractor weights, and inserting some parameter-efficient adaptation modules, lightweight enough to be trained in a few-shot manner. FLUTE~\cite{icml21_fsdg} manages it through selective fine-tuning of a tiny set of FILM~\cite{aaai18_film} parameters, while keeping most of them fixed. Despite this progress, the best way to manage the adaptation/generalisation trade-off in fine-tuning approaches to few-shot learning (FSL) is still an open question. For example, which layers should be fine-tuned? What kind of adapters should be inserted, and where? While PMF, TSA, ETT, FLUTE, and others provide some intuitive recommendations, we propose a more systematic approach to answer these questions.

In this paper, we advance the adaptation-based paradigm for FSL by developing a neural architecture search (NAS) algorithm to find the optimal adaptation architecture. Given an initial pre-trained feature extractor, our NAS determines the subset of the architecture that should be fine-tuned, as well as the subset of layers where adaptation modules should be inserted. We draw inspiration from recent work in NAS~\cite{iclr20_ofa,iccv21_autoformer,iccv21_fairnas,eccv20_spos,corr22_noah} that proposes revised versions of the stochastic Single-Path One-Shot (SPOS)~\cite{eccv20_spos} weight-sharing strategy. Specifically, given a pre-trained ResNet~\cite{cvpr2016_resnet} or Vision Transformer (ViT)~\cite{iclr21_vit}, we consider a search space defined by the inclusion or non-inclusion of task-specific adapters per layer, and the freezing or fine-tuning of learnable parameters per layer. Based on this search space, we construct a supernet~\cite{brock2018smash} that we train by sampling a random path in each forward pass~\cite{eccv20_spos}. Our supernet architecture is illustrated schematically in Figure~\ref{fig:supernet-architecture}, where the aforementioned decisions are drawn as decision nodes ($\diamond$), and possible paths are marked in dotted lines.

While the supernet training remains somewhat similar to the standard NAS approaches, the subsequent search poses new challenges due to the inherent characteristics of the FSL setting. Specifically, as cross-domain FSL considers a number of datasets including novel domains at test time, it becomes questionable whether searching for a single model -- which is the prevalent paradigm in NAS~\cite{cai2019proxylessnas,li2020sgas,liu2019darts,wang2021darts-pt} -- is the best choice. On the other hand, per-episode architecture selection is too slow and might overfit to the small support set. 

Motivated by these challenges, we propose a novel NAS algorithm that shortlists a small number of architecturally diverse configurations at training time, but defers the final selection until the dataset and episode is known at test time. We empirically show that this is not only computationally efficient, but also improves results noticeably, especially when only a limited amount of domains is available at training time. We term our method \longName{} (\shortName{}).

\shortName{} defines a generic search space that is relevant to both major architecture families (i.e., convolutional networks and transformers), and the choice of which specific adapter modules to consider is a hyperparameter, rather than a hard constraint. In this paper, we consider using adapter modules that are currently state-of-the-art for ResNets and ViTs (TSA and ETT, respectively), but more adaptation architectures can be added to the search space. 

Our contributions are summarised as follows: (i) We provide the first systematic Auto-ML approach to finding the optimal adaptation strategy that trades-off adaptation flexibility and overfitting risk in  multi-domain FSL. (ii) Our novel \shortName{} algorithm automatically determines which layers should be frozen or adapted, and where new adaptation parameters should be inserted for best few-shot adaptation. (iii) We advance the state-of-the-art in the well-established and challenging Meta-Dataset~\cite{iclr20_meta_dataset}, and the more recent and diverse Meta-Album~\cite{neurips22_meta_album} benchmarks.

\section{Related Work}
\label{sec:related_work}

\subsection{Adaptation for Few-shot Learning}
\noindent\textbf{Gradient-Based Adaptation}\quad
Parameter-efficient adaptation modules have been previously applied for multi-domain learning, and transfer learning. A seminal example of this are Residual Adapters~\cite{neurips17_residual_adapters}, which are lightweight 1x1 convolutional filters added to ResNet blocks. They were initially proposed for multi-domain learning, but are also useful for FSL, by providing the ability to update the feature extractor while being lightweight enough to avoid severe overfitting in the few-shot regime. Task-Specific Adapters (TSA)~\cite{cvpr22_tsa} use such adapters together with a URL~\cite{iccv21_url} pre-trained backbone to achieve state of the art results for CNNs on the Meta-Dataset benchmark~\cite{iclr20_meta_dataset}. Meanwhile, prompt~\cite{jia2022visual} and prefix~\cite{li2021prefix} tuning are established examples of parameter-efficient adaptation for transformer architectures for similar reasons. In FSL, Efficient Transformer Tuning (ETT)~\cite{tmlr22_ett} apply a similar strategy to few-shot ViT adaptation using a DINO~\cite{iccv21_dino} pre-trained backbone. 

PMF~\cite{cvpr22_pmf}, FLUTE~\cite{icml21_fsdg} and FT~\cite{dhillon2020baselineFSL} focus on adaptation of existing parameters without inserting new ones. To manage the adaptation/overfitting trade-off in the few-shot regime, PMF fine-tunes the whole ResNet or ViT backbone, but with carefully-managed learning rates. Meanwhile, FLUTE  hand-picks a set of FILM parameters with a modified ResNet backbone for few-shot fine-tuning, while keeping the majority of the feature extractor frozen. 

All of the methods above make heuristic choices about where to place adapters within the backbone, or for which parameters to allow/disallow fine-tuning. However, as different input layers represent different features~\cite{iccv21_autoformer,zeiler2014understandingCNN}, there is scope for making better decisions about which features to update. Furthermore, in the multi-domain setting different target datasets may benefit from different choices about which modules to update. This paper takes an Auto-ML NAS-based approach to systematically address this issue.

\noindent\textbf{Feed-Forward Adaptation}\quad
The aforementioned methods all use stochastic gradient descent to update the features during adaptation. We briefly mention CNAPS~\cite{neurips19_cnaps} and derivatives~\cite{cvpr20_simple_cnaps} as a competing line of work that use feed-forward networks to modulate the feature extraction process. However, these dynamic feature extractors are less able to generalise to completely novel domains than gradient-based methods~\cite{finn2018metaUniversal}, as the adaptation module itself suffers from an out of distribution problem.

\subsection{Neural architecture search}
Neural Architecture Search (NAS) is a large and well-studied topic~\cite{elsken2019nas} which we do not attempt to review in detail here. Mainstream NAS aims to discover new architectures that achieve high performance when training on a single dataset from scratch in a many-shot regime. To this end, research aims to develop faster search algorithms~\cite{abdelfattah2021zerocostNAS,eccv20_spos,liu2019darts,xiang2023zcpt}, and more effective search spaces~\cite{ci2021evolving,fang2019densely,radosavovic2019design_spaces,zhou2021autospace}. We build upon the popular SPOS~\cite{eccv20_spos} family of search strategies that encapsulate the entire search space inside a supernet that is trained by sampling paths randomly, and a search algorithm then determines the optimal path.

We develop an instantiation of the SPOS strategy for the multi-domain FSL problem. We construct a search space suited for parameter-efficient adaptation of a prior architecture to a new set of categories, and extend SPOS to learn on a suite of datasets, and efficiently generalise to novel datasets. This is different than the traditional SPOS paradigm of training and evaluating on the same dataset and same set of categories. 

While there exist some recent NAS works that try to address a similar ``train once, search many times'' problem efficiently~\cite{iclr20_ofa,li2020dna,malchanov2022lana,bert2021donna}, naively using these approaches has two serious shortcomings: i) They assume that after the initial supernet training, subsequent searches do not involve any training (e.g., a search is only performed to consider a different FLOPs constraint while accuracy of different configurations is assumed to stay the same) and thus can be done efficiently -- this is not true in the FSL setting as explained earlier. ii) Even if naively searching for each dataset at test time were computationally feasible, the few-shot nature of our setting poses a significant risk of overfitting the architecture to the small support set considered in each episode.
\begin{figure}[t]
  \centering
  \begin{subfigure}{0.97\linewidth}
    \includegraphics[width=\linewidth]{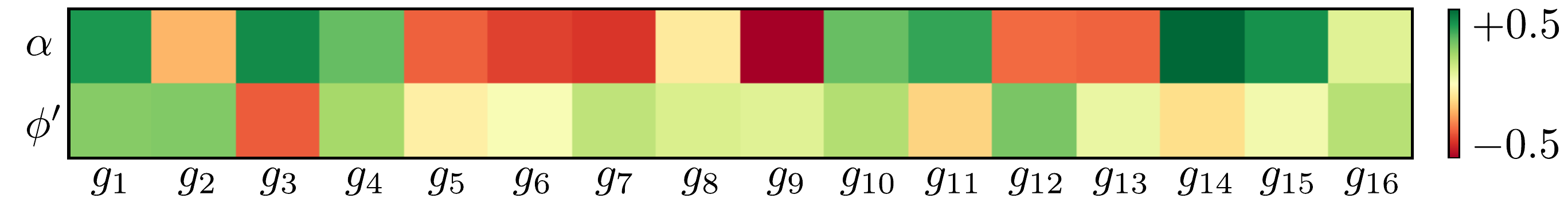}
    \caption{Correlation between inclusion/non-inclusion of learnable parameters $\alpha$ and $\phi'$, and validation performance.}
    \label{subfig:supernet-summary}
  \end{subfigure}
  \begin{subfigure}{0.97\linewidth}
    \includegraphics[width=\linewidth]{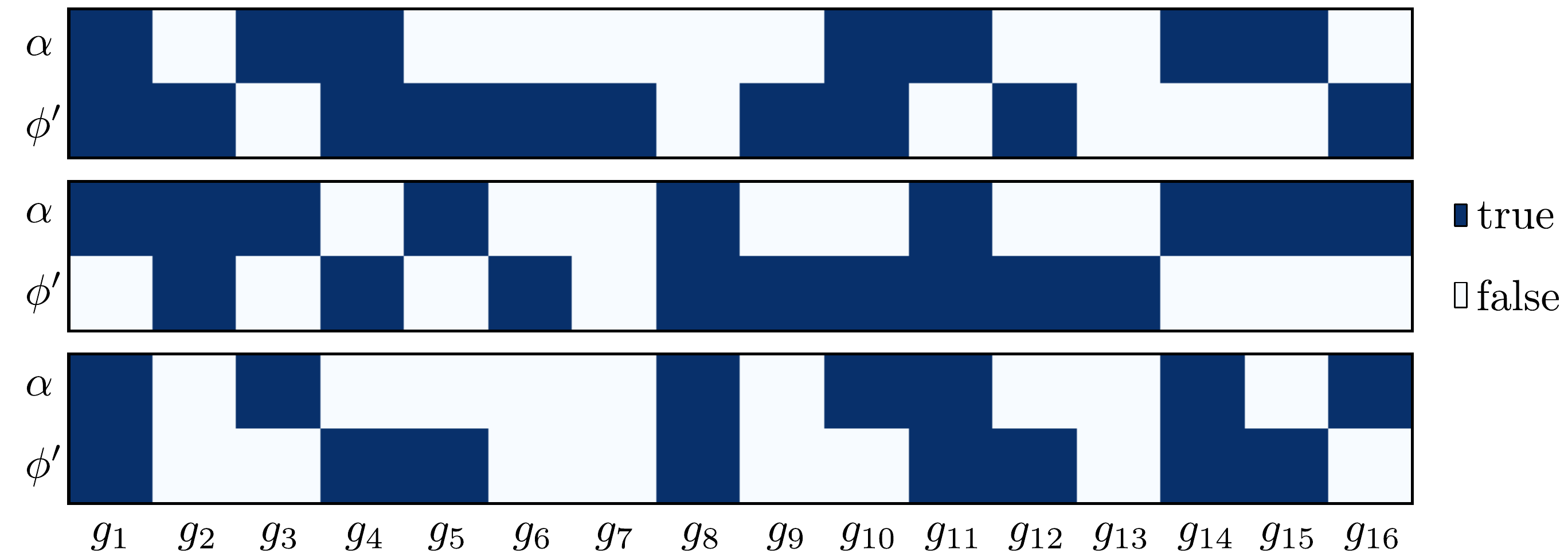}
    \caption{Top 3 performing paths subject to diversity constraint.}
    \label{subfig:supernet-top3-diverse}
  \end{subfigure}
  \caption{Qualitative analysis of our architecture search. Fig.~\ref{subfig:supernet-summary} summarises the whole search space by answering the question: \emph{How important is to adapt ($\alpha$) or fine-tune ($\phi'$) each block?} The color of each square indicates the point-biserial correlation (over all searched architectures) between adapting/fine-tuning layer $g_i$ and validation performance.
  Fig.~\ref{subfig:supernet-top3-diverse} shows the top 3 performing candidates subject to a diversity constraint, after 15 generations of evolutionary search. Dark blue indicates that the layer is adapted/fine-tuned and light blue that it is not.}
  \label{fig:supernet-qualitative}
  \vspace{-1em}
\end{figure}

\begin{figure*}[t]
  \centering
  \includegraphics[width=\linewidth]{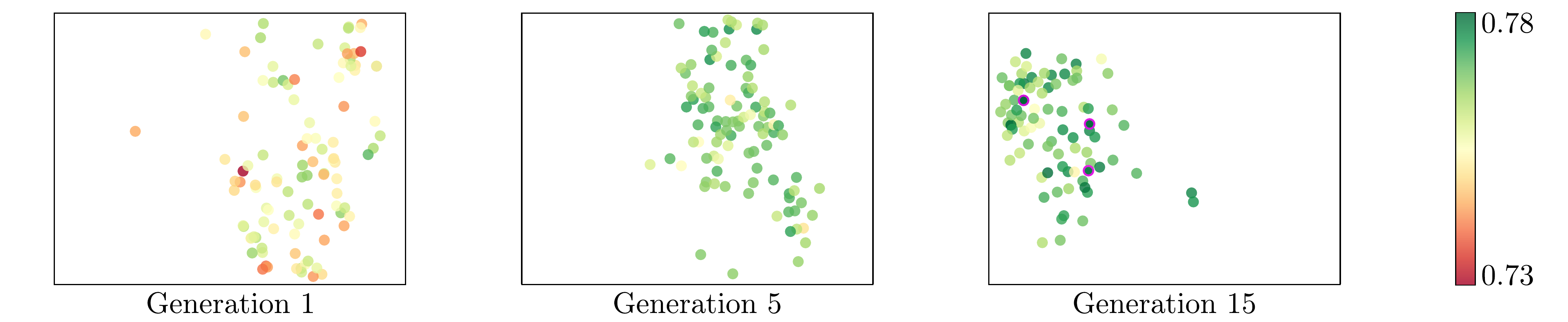}
  \caption{Population of paths(candidate architectures) in the search space after 1, 5, and 15 generations of evolutionary search. Each dot is a 2-d TSNE projection of the binary vector representing an architecture, and its color shows the validation performance for that architecture. The supernet contains a wide variety of models in terms of validation performance, and the search algorithm converges to a well-performing population. The top 3 performing paths that are given in~\ref{subfig:supernet-top3-diverse} are highlighted in the far right figure (Generation 15) in purple outline.}
  \label{fig:search-qualitative}
  \vspace{-1em}
\end{figure*}

\section{Neural Fine-Tuning Search}
\label{sec:method}

\subsection{Few-Shot Learning Background}

Let $\mathcal{D}=\{\mathcal{D}_i\}_{i=1}^D$ be the set of $D$ classification domains, and $\bar{\mathcal{D}}=\{X, Y\} \in \mathcal{D}$ a task containing $n$ samples along with their designated true labels $\{\bar{X}, \bar{Y}\}=\{x_j,y_j\}_{j=1}^{n}$. Few-shot classification is defined as the problem of learning to correctly classify a query set $\mathcal{Q}=\{X_\mathcal{Q}, Y_\mathcal{Q}\} \sim \bar{\mathcal{D}}$ by training on a support set $\mathcal{S}=\{X_\mathcal{S}, Y_\mathcal{S}\} \sim \bar{\mathcal{D}}$ that contains very few examples. This can be achieved by finding the parameters $\theta$ of a classifier $f_\theta$ with the objective
\begin{equation}
    \argmax_\theta\prod_{\mathcal{D}} p(Y_\mathcal{Q} | f_\theta(\mathcal{S}, X_\mathcal{Q})).
    \label{eq:fsl-objective}
\end{equation}
In practice, if $\theta$ is randomly initialised and trained using stochastic gradient descent on a small support set $\mathcal{S}$, it will overfit and fail to generalise to $\mathcal{Q}$. To address this issue, one can exploit knowledge transfer from some seen classes to the novel classes. Formally, each domain $\bar{\mathcal{D}}$ is partitioned into two disjoint sets $\bar{\mathcal{D}}_\text{train}$ and $\bar{\mathcal{D}}_\text{test}$, which are commonly referred to as ``meta-train'' and ``meta-test'', respectively. The labels in these sets are also disjoint, i.e., $Y_\text{train} \cap Y_\text{test}=\emptyset$. In that case, $\theta$ is trained by maximising the objective in Eq.~\ref{eq:fsl-objective} using the meta-train set, but the overall objective is to perform adequately when transferring knowledge to meta-test.

The knowledge transferred from meta-train to meta-test can take various forms~\cite{hospedales20201metaSurveyPAMI}. As discussed earlier, we aim to generalise a family of few-shot methods~\cite{cvpr22_pmf,cvpr22_tsa,tmlr22_ett} where parameters $\theta$ are transferred before a subset of them $\phi\subset\theta$ are fine-tuned; and possibly extended by attaching additional ``adapter'' parameters $\alpha$ that are trained for the target task.
For meta-test, Eq.~\ref{eq:fsl-objective} can therefore be rewritten as
\begin{equation}
    \argmax_{\alpha,\phi}\prod_{\mathcal{D_{\text{test}}}} p(Y_\mathcal{Q} | f_{\alpha,\phi}(\mathcal{S}, X_\mathcal{Q})),
    \label{eq:fsl-objective-nfts}
\end{equation}
In this paper, we focus on the problem of finding the optimal adaptation strategy in terms of (i) the optimal subset of parameters $\phi \subset \theta$ that need to be fine-tuned, and (ii) the optimal task-specific parameters $\alpha$ to add.

\begin{table}[b]
  \centering
  \resizebox{1.\linewidth}{!}{
    \begin{tabular}{l|c|c}
    \toprule
    & $g_{\phi,\phi',\alpha}(x)$ (ResNet) & $g_{\phi,\phi',\alpha}(x)$ (ViT) \\
    \midrule
    $\phi\phantom{'}$, $-$ & $ g_\phi(x)$ & $z(A_{qkv}[q\;;\;g_{\phi}(x)])$ \\
    $\phi\phantom{'}$, $\alpha$ & $g_{\phi}(x) + h_\alpha(x)$ & $z(A_{qkv}[q\;;\;g_{\phi}(x)] + h_{\alpha1}) + h_{\alpha2}$ \\
    $\phi'$, $-$ & $g_{\phi'}(x)$ & $z(A_{qkv}[q\;;\;g_{\phi'}(x)])$ \\
    $\phi', \alpha$ & $g_{\phi'}(x) + h_\alpha(x)$ & $z(A_{qkv}[q\;;\;g_{\phi'}(x)] + h_{\alpha1}) + h_{\alpha2}$ \\
    \bottomrule
    \end{tabular}
  }
  \caption{The search space, as described in Section ~\ref{sec:search-space}. When sampling a layer $g_{\phi,\phi',\alpha}$, it can be sampled in one of the following variants: (i) $\phi$: fixed pre-trained parameters, no adaptation, (ii) $\alpha$: fixed pre-trained parameters, with adaptation, (iii) $\phi'$: fine-tuned parameters, no adaptation, (iv) $\phi', \alpha$ fine-tuned-parameters, with adaptation.}
  \label{tab:search-space}
\end{table}

\subsection{Defining the search space}
\label{sec:search-space}

Let $g_{\phi_k}$ be the minimal unit for adaptation in an architecture. We consider these to be the repeated units in contemporary deep architectures, e.g., a convolutional layer in a ResNet, or a self-attention block in a ViT. If the feature extractor $f_\theta$ comprises of $K$ such units with learnable parameters $\phi_k$, then we denote $\theta=\bigcup_{k=1}^K\phi_k$, assuming all other parameters are kept fixed. For brevity in notation we will now omit the indices and refer to every such layer as $g_\phi$. Following the state-of-the-art~\cite{cvpr22_pmf,cvpr22_tsa,icml21_fsdg,tmlr22_ett}, let us also assume that task-specific adaptation can be performed either by inserting additional adapter parameters $\alpha$ into $g_\phi$, or by fine-tuning the layer parameters $\phi$.

This allows us to define the search space as two independent binary decisions per layer: (i) The inclusion or non-inclusion of an adapter module attached to $g_\phi$, and (ii) the decision of whether to use the pre-trained parameters $\phi$, or replace them with their fine-tuned counterparts $\phi'$. The size of the search space is, therefore, $(2^2)^K=4^K$. For ResNets, we use the proposed adaptation architecture of TSA~\cite{cvpr22_tsa}, where a residual adapter $h_\alpha$, parameterised by $\alpha$, is connected to $g_\phi$
\begin{equation}
    g_{\phi,\phi',\alpha}(x) = g_{\phi,\phi'}(x) + h_\alpha(x),
\end{equation}
where $x\in\mathbb{R}^{W,H,C}$. For ViTs, we use the proposed adaptation architecture of ETT~\cite{tmlr22_ett}, where a tuneable prefix is prepended to the multi-head self-attention module $A_{qkv}$, and a residual adapter is appended to both $A_{qkv}$ and the feed-forward module $z$ in each decoder block
\begin{equation}
    g_{\phi,\phi',\alpha}(x) = z(A_{qkv}[q\;;\;g_{\phi,\phi'}(x)] + h_{\alpha1}) + h_{\alpha2},
\end{equation}
where $x\in\mathbb{R}^{D}$ and $[\cdot\;;\;\cdot]$ denotes the concatenation operation. Note that in the case of ViTs the adapter is not a function of the input features, but simply an added offset.

Irrespective of the architecture, every layer $g_{\phi,\phi',\alpha}$ is parameterised by three sets of parameters, $\phi$, $\phi'$, and $\alpha$, denoting the initial parameters, fine-tuned parameters and adapter parameters respectively. Consequently, when sampling a configuration (i.e., path) from that search space, every such layer can be sampled as one of the variants listed in Table~\ref{tab:search-space}.

\subsection{Training the supernet}
\label{sec:supernet-training}

\begin{algorithm}[t]
\DontPrintSemicolon
\SetKwInput{KwModel}{Model}
\SetKwRepeat{Do}{do}{while}
\KwIn{Supernet $f_{\theta,\alpha,\phi'}$. Datasets $\mathcal{D}$. Step sizes $\eta_1$, $\eta_2$. Path pool $P$. Prototypical loss $\mathcal{L}$ (Eq.~\ref{eq:proto-loss}).}
\KwOut{Trained supernet $f_{\theta,\alpha,\phi'}$.}
\Repeat{\normalfont{prototypical loss converges}} {
    Sample dataset $\bar{\mathcal{D}} \sim \mathcal{D}$\;
    Sample episode $\mathcal{S}$, $\mathcal{Q}$ $\sim \bar{\mathcal{D}}$\;
    Sample path $p \sim P$ with learnable parameters $\alpha_p$, $\phi_p'$ and frozen parameters $\phi_p \subset \theta$\;
    $\alpha_p \longleftarrow \alpha_p - \eta_1 \nabla_{\alpha_p} \mathcal{L}(f_{\theta,\alpha,\phi'}^p, \mathcal{S}, \mathcal{Q})$\;
    $\phi_p' \longleftarrow \phi'_p - \eta_2 \nabla_{\phi_p'} \mathcal{L}(f_{\theta,\alpha,\phi'}^p, \mathcal{S}, \mathcal{Q})$\;
}
\caption{Supernet training.}
\label{algo:supernet-training}
\end{algorithm}

Following SPOS~\cite{eccv20_spos}, our search space is actualised in the form of a supernet $f_{\theta,\alpha,\phi'}$; a ``super'' architecture that contains all possible architectures derived from the decisions detailed in Section~\ref{sec:search-space}. It is parameterised by: (i) $\theta$, the frozen parameters from the backbone architecture $f_\theta$, (ii) $\alpha$, from the adapters $h_\alpha$, and (iii) $\phi'$, from the fine-tuned parameters per layer $g_{\phi,\phi',\alpha}$.

We use a prototypical loss $\mathcal{L}(f, S, Q)$ as the core objective during supernet training and the subsequent search and fine-tuning.
\begin{equation}
    \mathcal{L}(f, \mathcal{S}, \mathcal{Q}) = 
    \frac{1}{|\mathcal{Q}|} \sum_{i=1}^{|\mathcal{Q}|} \log
    \frac{
        e^{-d_{cos}(C_{\mathcal{Q}_i}, f(\mathcal{Q}_i))}
    }{
        \sum_{j=1}^{|C|} e^{-d_{cos}(C_j, f(\mathcal{Q}_i))}
    }
    ,
    \label{eq:proto-loss}
\end{equation}
where $C_{\mathcal{Q}_i}$ denotes the embedding of the class centroid that corresponds to the true class of $\mathcal{Q}_i$, and $d_{cos}$ denotes the cosine distance. The set of class centroids $C$ is computed as the mean embeddings of support examples that belong to the same class:
\begin{equation}
    C = \Bigl\{ \frac{1}{|\mathcal{S}^{y=l}|} \sum_{i=1}^{|\mathcal{S}|} f(\mathcal{S}^{y=l}_i) \Bigl\}_{l=1}^{L},
\end{equation}
where $L$ denotes the number of unique labels in $\mathcal{S}$.

For supernet training specifically, let $P$ be a set of size $4^K$, enumerating all possible sequences of $K$ layers that can be sampled from the search space.
Denoting a path sampled from the supernet as $f_{\theta,\alpha,\phi'}^p$, we minimise an expectation of the loss in Eq.~\ref{eq:proto-loss} over multiple episodes and paths, so the final objective becomes:
\begin{equation}
    \argmin_{\alpha,\phi'} \; \mathbb{E}_{p\sim P} \mathbb{E}_{\mathcal{S},\mathcal{Q}} \; \mathcal{L}(f_{\theta,\alpha,\phi'}^p, \mathcal{S}, \mathcal{Q}).
\end{equation}
 Algorithm~\ref{algo:supernet-training} summarises the supernet training algorithm in pseudocode.

\subsection{Searching for an optimal path}
\label{sec:method-search}

\begin{algorithm}[t]
\DontPrintSemicolon
\SetKwInput{KwModel}{Model}
\SetKwRepeat{Do}{do}{while}
\KwIn{Supernet $f_{\theta,\alpha,\phi'}$. Datasets $\mathcal{D}$. Step sizes $\eta_1$, $\eta_2$. Prototypical loss $\mathcal{L}$ (Eq.~\ref{eq:proto-loss}). NCC accuracy $A$ (Eq.~\ref{eq:ncc-acc}).}
\KwOut{Optimal path $p^*$.}
Initialise population $P$ randomly\;
Initialise fitness of $P$ as $\Psi_P \longleftarrow 0$\;
\Repeat{\normalfont{population fitness converges or max. iterations}} {
    Sample episodes from all datasets $\mathcal{S}$, $\mathcal{Q}$ $\sim \mathcal{D}$\;
    \For{\normalfont{each candidate $p \in P$}}{
        \For{\normalfont{a small number of epochs}} {
            $\alpha_p \longleftarrow \alpha_p - \eta_1 \nabla_{\alpha_p} \mathcal{L}(f_{\theta,\alpha,\phi'}^p, \mathcal{S}, \mathcal{S})$\;        
            $\phi_p' \longleftarrow \phi_p' - \eta_2 \nabla_{\phi_p'} \mathcal{L}(f_{\theta,\alpha,\phi'}^p, \mathcal{S}, \mathcal{S})$\;
        }        
        $\Psi_p$ $\longleftarrow$ $A(f_{\theta,\alpha,\phi'}^p, \mathcal{S}, \mathcal{Q})$\;
    }
    offspring $\longleftarrow$ recombine the $M$ best candidates of $P$ w.r.t. $\Psi_P$\;
    $P$ $\longleftarrow$ $P$ + offspring\;
    eliminate the $M$ worst candidates of $P$ w.r.t. $\Psi_P$\;
}
\caption{Training time evolutionary search.}
\label{algo:search}
\end{algorithm}

A supernet $f_{\theta,\alpha,\phi'}$ trained with the method described in Section~\ref{sec:supernet-training} contains $4^K$ models, intertwined via weight sharing. As explained in Section~\ref{sec:introduction}, our goal is to search for the best-performing one, but the main challenge is related to the fact that we do not know what data is going to be used for adaptation at test time. One extreme approach, would be to search for a single solution at training time and simply use it throughout the entire test, regardless of the potential domain shift. Another, would be to defer the search and perform it from scratch each time a new support set is given to us at test time.
However, both have their shortcomings. As such, we propose a generalization of this process where searching is split into two phases -- one during training, and a subsequent one during testing.

\noindent\textbf{Meta-training time.}\quad
The search is responsible for pre-selecting a set of $N$ models from the entire search space. Its main purpose is to mitigate potential overfitting that can happen at test time, when only a small amount of data is available, while providing enough diversity to successfully adjust the architecture to the diverse set of test domains. Formally, we search for a sequence of paths $(p_1, p_2, ..., p_N)$ where:

\begin{gather}
\label{eq:search_obj_train}
p_k = \argmax_{p \in P} \mathbb{E}_{\mathcal{S},\mathcal{Q}} A(f_{\theta,\alpha^*,\phi'^*}^p, \mathcal{S},\mathcal{Q}), \quad \text{s.t.} \\
\label{eq:fine_tuning}
\alpha^*,\phi'^* = \argmin_{\alpha, \phi'} \mathcal{L}(f_{\theta,\alpha,\phi'}^p, \mathcal{S}, \mathcal{S}) \\ \label{eq:diversity_constraint}
\forall_{j=1,...,k-1} \;\; d_{cos}(p_k, p_j) \geq T,
\end{gather}
where $T$ denotes a scalar threshold for the cosine distance between paths $p_k$ and $p_j$, and $A$ is the classification accuracy of a nearest centroid classifier (NCC)~\cite{neurips17_protonet},
\begin{equation}
   A(f, \mathcal{S}, \mathcal{Q}) = \frac{1}{|\mathcal{Q}|} \sum_{i=1}^{|\mathcal{Q}|}[\argmin_j d_{cos} ( C_{\mathcal{Q}_j}, f(\mathcal{Q}_i) ) = Y_{\mathcal{Q}_i}].
   \label{eq:ncc-acc}
\end{equation}

Noticeably, we measure accuracy of a solution using a query set, after fine-tuning on a separate support set (Eq.~\ref{eq:fine_tuning}), then average across multiple episodes to avoid overfitting to a particular support set (Eq.~\ref{eq:search_obj_train}).
We also employ a diversity constraint, in the form of cosine distance between binary encodings of selected paths (Eq.~\ref{eq:diversity_constraint}), to allow for sufficient flexibility in the following test time search.

To efficiently obtain sequence $\{ p_  1,...,p_N \}$, we use evolutionary search to find points that maximise Eq.~\ref{eq:search_obj_train}, and afterwards select the $N$ best performers from the evolutionary search history that satisfy the constraint in Eq.~\ref{eq:diversity_constraint}.
Algorithm~\ref{algo:search} summarises training-time search.

\noindent\textbf{Meta-testing time.}\quad
For a given meta-test episode, we decide which one of the pre-selected $N$ models is best suited for adaptation on the given support set data. 
It acts as a failsafe to counteract the bias of the initial selection made at training time in cases when the support set might be particularly out-of-domain.
Formally, the final path $p^*$ to be used in a particular episode is defined as:
\begin{gather}
\label{eq:search_obj_test}
p^* = \argmin_{p \in \{ p_  1,...,p_N \}} \mathcal{L}(f_{\theta, \alpha^*, \phi'^*}^p, \mathcal{S}, \mathcal{S}),\quad\text{s.t.} \\\label{eq:fine_tuning_test}
\alpha^*, \phi'^* = \argmin_{\alpha, \phi'} \mathcal{L}(f_{\theta, \alpha, \phi'}^p, \mathcal{S}, \mathcal{S})
\end{gather}

Noticeably, we test each of the $N$ models by fine-tuning it on the support set (Eq.~\ref{eq:fine_tuning_test}) and testing its performance on the same support set (Eq.~\ref{eq:search_obj_test}). This is because the support set is the only source of data we have at test time and we cannot extract a disjoint validation set from it without risking the quality of the fine-tuning process. It is important to note that, while this step risks overfitting, the pre-selection of models at training time, as described previously, should already limit the subsequent search to only models that are unlikely to overfit. Since $N$ is kept small in our experiments, we use a naive grid search to find $p^*$.

This approach is a generalization of the existing NAS approaches, as it recovers both when $N=1$ or $N=4^K$. Our claim is that intermediate values of $N$ are more likely to give us better results than any of the extremes, due to the reasons mentioned earlier. In particular, we would expect pre-selecting $1 < N \ll 4^K$ models to introduce reasonable overhead at test time while improving results, especially in cases when exposure to different domains might be limited at training time. In our evaluation we compare $N=3$ and $N=1$ to test this hypothesis. We do not include comparison to $N=4^K$ as it is computationally infeasible  in our setting (performing equivalent of training time search for each test episode would require us to fine-tune $\approx 14*10^6$ models in total).

\begin{table*}[t]
  \centering
  \resizebox{1.\linewidth}{!}{
    \begin{tabular}{cl|ccccccccccccc|c}
    \toprule
    & Method & Aircrafts & Birds & DTD & Fungi & ImageNet & Omniglot & QuickDraw & Flowers & CIFAR-10 & CIFAR-100 & MNIST & MSCOCO & Tr. Signs & Average \\
    \toprule
      \multirow{6}{*}{\rotatebox{90} {ResNet-18}}
    & FLUTE \cite{iclr21_urt} & 48.5 & 47.9 & 63.8 & 31.8 & 46.9 & 61.6 & 57.5 & 80.1 & 65.4 & 52.7 & 80.8 & 41.4 & 46.5 & 52.6 \\
    & ProtoNet \cite{neurips17_protonet} & 53.1 & 68.8 & 66.6 & 39.7 & 50.5 & 60.0 & 49.0 & 85.3 & -    &    - &    - & 41.0 & 47.1 & 56.1 \\
    & BOHB \cite{saikia2020optimized} & 54.1 & 70.7 & 68.3 & 41.4 & 51.9 & 67.6 & 50.3 & 87.3 & -    &    - &    - & 48.0 & 51.8 & 59.2 \\
    & FO-MAML \cite{iclr20_meta_dataset} & 63.4 & 69.8 & 70.8 & 41.5 & 52.8 & 61.9 & 59.2 & 86.0 & -    &    - &    - & 48.1 & 60.8 & 61.4 \\
    & TSA \cite{cvpr22_tsa} & 72.2 & 74.9 & 77.3 & 44.7 & 59.5 & 78.2 & \textbf{67.6} & 90.9 & 82.1 & 70.7 & 93.9 & 59.0 & \textbf{82.5} & 73.3 \\
    & \shortName{} & \textbf{74.9} & \textbf{76.5} & \textbf{81.6} & \textbf{50.5} & \textbf{62.7} & \textbf{80.2} & 67.2 & \textbf{94.5} & \textbf{83.0} & \textbf{71.5} & \textbf{94.0} & \textbf{59.7} & 81.9 & \textbf{75.2} \\
    \midrule
    \multirow{3}{*}{\rotatebox{90} {ViT-S}} 
    & PMF$^*$ \cite{cvpr22_pmf} & 76.8 & 85.0 & 86.6 & 54.8 & \textbf{74.7} & 80.7 & 71.3 & 94.6 & -    &    -    & - & \textbf{62.6} & \textbf{88.3} & 77.5 \\
    & ETT \cite{tmlr22_ett} & 79.9 & \textbf{85.9} & \textbf{87.6} & 61.8 & 67.4 & 78.1 & 71.3 & \textbf{96.6} & -    &    -    & - & 62.3 & 85.1 & 77.6 \\
    & \shortName{} & \textbf{83.0} & 85.5 & \textbf{87.6} & \textbf{62.2} & 71.0 & \textbf{81.9} & \textbf{74.5} & 96.0 & 79.4 & 72.6 & 95.2 & \textbf{62.6} & 87.9 & \textbf{79.2} \\
    \bottomrule
    \end{tabular}
    \vspace{-1em}
  }
  \caption{Comparison to the state-of-the art methods on Meta-Dataset. Single domain setting -- only ImageNet is seen during training and search. Reporting mean accuracy over 600 episodes.  $^*$ Additional data used for training.}
  \label{tab:sota-single-domain}
\end{table*}

\begin{table*}[t]
  \centering
  \resizebox{1.\linewidth}{!}{
    \begin{tabular}{cl|ccccccccccccc|c}
    \toprule
    &Method & Aircrafts & Birds & DTD & Fungi & ImageNet & Omniglot & QuickDraw & Flowers & CIFAR-10 & CIFAR-100 & MNIST & MSCOCO & Tr. Signs & Average \\
    \toprule
    \multirow{7}{*}{\rotatebox{90} {ResNet-18}} 
    & CNAPS \cite{neurips19_cnaps} & 83.7 & 73.6 & 59.5 & 50.2 & 50.8 & 91.7 & 74.7 & 88.9 &    - &    - &    - & 39.4 & 56.5 & 66.9 \\
    & SCNAPS \cite{wang2019simpleshot} & 82.0 & 74.8 & 68.8 & 46.6 & 58.4 & 91.6 & 76.5 & 90.5 & 74.9 & 61.3 & 94.6 & 48.9 & 57.2 & 69.5 \\
    & SUR \cite{dvornik2020selecting} & 85.5 & 71.0 & 71.0 & 64.3 & 56.2 & 94.1 & 81.8 & 82.9 & 66.5 & 56.9 & 94.3 & 52.0 & 51.0 & 71.4 \\
    & URT \cite{iclr21_urt} & 85.8 & 76.2 & 71.6 & 64.0 & 56.8 & 94.2 & 82.4 & 87.9 & 67.0 & 57.3 & 90.6 & 51.5 & 48.2 & 71.8 \\
    & FLUTE \cite{icml21_fsdg} & 82.8 & 75.3 & 71.2 & 48.5 & 58.6 & 92.0 & 77.3 & 90.5 & 75.4 & 62.0 & 96.2 & 52.8 & 63.0 & 72.7 \\
    & URL \cite{iccv21_url} & 89.4 & 80.7 & 77.2 & 68.1 & 58.8 & 94.5 & 82.5 & 92.0 & 74.2 & 63.5 & 94.7 & 57.3 & 63.3 & 76.6 \\
    & TSA \cite{cvpr22_tsa} & 89.9 & 81.1 & 77.5 & 66.3 & 59.5 & \textbf{94.9} & 81.7 & \textbf{92.2} & 82.9 & 70.4 & \textbf{96.7} & 57.6 & \textbf{82.8} & 78.4 \\
    & \shortName{} & \textbf{90.1} & \textbf{83.8} & \textbf{82.3} & \textbf{68.4} & \textbf{61.4} & 94.3 & \textbf{82.6} & \textbf{92.2} & \textbf{83.0} & \textbf{75.1} & 95.4 & \textbf{58.8} & 81.9 & \textbf{80.7} \\
    \midrule
    \multirow{3}{*}{\rotatebox{90} {ViT-S\hspace{-0.3cm}}} 
    & PMF$^*$ \cite{cvpr22_pmf} & 88.3 & 91.0 & \textbf{86.6} & 74.2 & \textbf{74.6} & 91.8 & 79.2 & \textbf{94.1} & -    & -    & -    & 62.6 & \textbf{88.9} & 83.1 \\
    & \shortName{} & \textbf{89.1} & \textbf{92.5} & 86.3 & \textbf{75.1} & \textbf{74.6} & \textbf{92.0} & \textbf{80.6} & 93.5 & 75.9 & 70.8 & 91.3 & \textbf{62.8} & 87.2 & \textbf{83.4} \\
    \bottomrule
    \end{tabular}
  }
  \caption{Comparison to the state-of-the art methods on Meta-Dataset. Multi-domain setting -- the first 8 datasets are seen during training and search. Reporting mean accuracy over 600 episodes. $^*$ Additional data used for training.}
  \label{tab:sota-multi-domain}
  \vspace{-1em}
\end{table*}

\section{Experiments}
\label{sec:experiments}

\subsection{Experimental setup}

\noindent\textbf{Evaluation on Meta-Dataset}\quad
We evaluate \shortName{} on the extended version of Meta-Dataset~\cite{neurips19_cnaps,iclr20_meta_dataset}, currently the most commonly used benchmark for few-shot classification, consisting of 13 publicly available datasets: FGVC Aircraft, CU Birds, Describable Textures (DTD), FGVCx Fungi, ImageNet, Omniglot, QuickDraw, VGG Flowers, CIFAR-10/100, MNIST, MSCOCO, and Traffic Signs. There are 2 evaluation protocols: single domain learning and multi-domain learning. In the single domain setting, only ImageNet is seen during training and meta-training, while in the multi-domain setting the first eight datasets are seen (FGVC Aircraft to VGG Flower). For meta-testing at least 600 episodes are sampled for each domain.

\noindent\textbf{Evaluation on Meta-Album}\quad
Further, we evaluate \shortName{} on the more recently introduced Meta-Album~\cite{neurips22_meta_album}. Meta-Album is more diverse than Meta-Dataset. We use the currently available Sets 0-2, which contain over 1000 unique labels across 30 datasets spanning 10 domains including microscopy, remote sensing, manufacturing, plant disease, character recognition and human action recognition tasks, etc. Unlike Meta-Dataset, where their default evaluation protocol is variable-way variable-shot, Meta-Album evaluation follows a 5-way variable-shot setting, where the number of shots is typically 1, 5, 10 and 20. For meta-testing, results are averaged over 1800 episodes.

\noindent\textbf{Architectures}\quad
We employ two different backbone architectures, a ResNet-18~\cite{cvpr2016_resnet} and a ViT-small~\cite{iclr21_vit}. Following TSA~\cite{cvpr22_tsa}, the ResNet-18 backbone is pre-trained on the seen domains with the knowledge-distillation method URL~\cite{iccv21_url} and, following ETT~\cite{tmlr22_ett}, the ViT-small backbone is pre-trained on the seen portion of ImageNet with the self-supervised method DINO~\cite{iccv21_dino}. We consider TSA residual adapters~\cite{cvpr22_tsa,neurips17_residual_adapters} for ResNet and Prefix Tuning~\cite{li2021prefix,tmlr22_ett} adapters for ViT. This is mainly to enable direct comparison with prior work on the same base architectures that use exactly these same adapter families, without introducing new confounders in terms of mixing adapter types \cite{cvpr22_tsa,tmlr22_ett}. However our framework is flexible, meaning it can accept any adapter type, or even multiple types in its search space.

\subsection{Comparison to state-of-the-art}

\noindent\textbf{Meta-Dataset}\quad
The results on Meta-Dataset are shown in Table~\ref{tab:sota-single-domain} and Table~\ref{tab:sota-multi-domain} for single-domain and multi-domain training setting respectively. We can see that \shortName{} obtains the best average performance across all the competitor methods for both ResNet and ViT architectures. The margins over prior state-of-the-art are often substantial for this benchmark with +1.9\% over TSA in ResNet-18 single domain, +2.3\% in multi-domain and +1.6\% over ETT in VIT-small single domain. The increased margin in the multi-domain case is intuitive, as our framework has more data with which to learn the optimal path(s). 

We re-iterate that PMF, ETT, and TSA are special cases of our search space corresponding respectively to: (i) Fine-tune all and include no adapters, (ii) Include ETT adapters at every layer while freezing all backbone weights and (iii) Include TSA adapters at every layer while freezing all backbone weights. We also share initial pre-trained backbones with ETT and TSA (but not PMF, as it uses a stronger pre-trained model with additional data). Thus the margins achieved over these competitors are attributable to our systematic approach to finding suitable architectures in terms of where to fine-tune and where to insert new adapter parameters. 

\begin{figure}[t]
  \centering
  \includegraphics[width=0.8\linewidth]{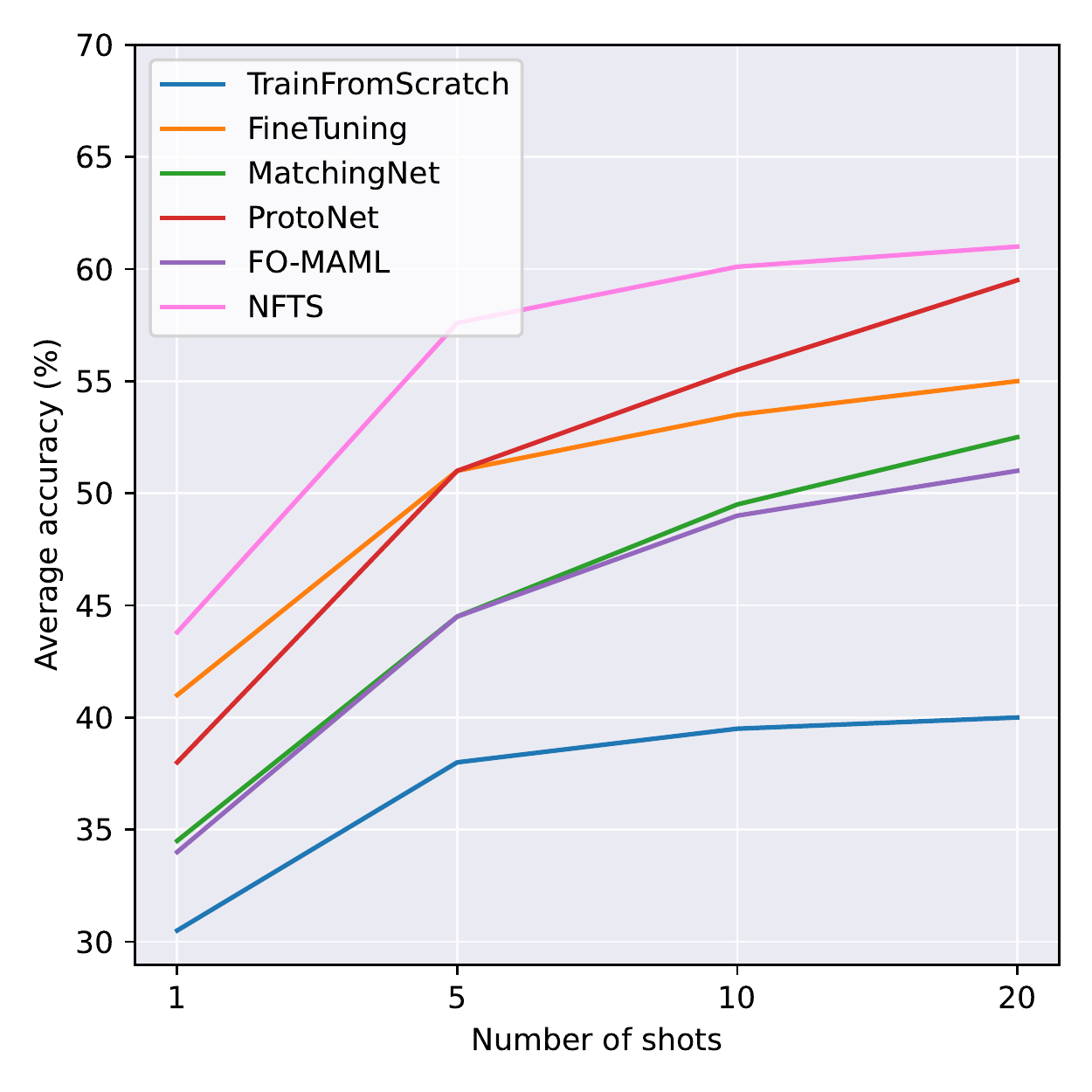}
  \caption{Comparison of our method against Meta-Album baselines, as reported in Fig. 2 of their paper~\cite{neurips22_meta_album}. The setting is 5-way [1, 5, 10, 20]-shot, and accuracy scores are averaged over 1800 tasks drawn from Set0, Set1 and Set2.}
  \label{fig:meta-album-results}
  \vspace{-1em}
\end{figure}

\noindent\textbf{Meta-Album}\quad
The results on Meta-Album are shown in Figure~\ref{fig:meta-album-results} as a function of number of shots within the 5-way setting, following~\cite{neurips22_meta_album}. We can see that across the whole range of support set sizes, our \shortName{} dominates all of the well-tuned baselines from~\cite{neurips22_meta_album}. The margins are substantial, greater than 5\% at 5-way/5-shot operating point, for example. This result confirms that our framework scales to even more diverse datasets and domains than those considered previously in Meta-Dataset. 

\subsection{Ablation study}\label{sec:ablation}

\begin{table}[t]
  \centering
  \resizebox{0.8\linewidth}{!}{
    \begin{tabular}{cl|cc}
    \toprule
    &Method & Single Domain & Multi-Domain \\
    \toprule
    \multirow{6}{*}{\rotatebox{90} {ResNet-18}} & $\phi\phantom{'},-$ & 67.8 & 67.8 \\
    & $\phi\phantom{'},\alpha$ & 70.4 & 76.5 \\
    & $\phi',-$ & 70.2 & 76.3 \\
    & $\phi', \alpha$ & 70.8 & 76.9 \\
    &\shortName{}-1 & 73.6 & 80.1 \\
    &\shortName{}-N & 75.2 & 80.7 \\
    \midrule
    \multirow{6}{*}{\rotatebox{90} {ViT-S}} & $\phi\phantom{'},-$ & 71.8 & 71.8 \\
    & $\phi\phantom{'},\alpha$ & 73.8 & 77.3 \\
    & $\phi',-$ & 74.0 & 77.5 \\
    & $\phi', \alpha$ & 74.4 & 78.9 \\
    &\shortName{}-1 & 78.7 & 83.1 \\
    &\shortName{}-N & 79.2 & 83.4 \\
    \bottomrule
    \end{tabular}
  }
  \caption{Ablation study on Meta-Dataset comparing four special cases of the search space in terms of average accuracy: (i) $\phi,-$: No adaptation, no fine-tuning, (ii) $\phi,\alpha$: Adapt all, (iii) $\phi',-$: Fine-tune all, (iv) $\phi', \alpha$: Adapt and fine-tune all. \shortName{}-\{1,N\} refer to conventional and deferred episode-wise NAS respectively.
  }
  \label{tab:ablation-short}
  \vspace{-1em}
\end{table}

To analyse more precisely the role that our architecture search plays in few-shot performance, we also conduct an ablation study of our final model against four corners of our search space: (i) Initial model only, using a pre-trained feature extractor and simple NCC classifier, which loosely corresponds to SimpleShot~\cite{wang2019simpleshot}, (ii) Full adaptation only, using a fixed feature extractor, which loosely corresponds to TSA~\cite{cvpr22_tsa}, ETT~\cite{tmlr22_ett}, FLUTE~\cite{icml21_fsdg}, and others -- depending on base architecture and choice of adapter, (iii) Fully fine-tuned model, which loosely corresponds to PMF~\cite{cvpr22_pmf}, and (iv) Combination of full fine-tuning and adaptation. From the results in Table~\ref{tab:ablation-short} we can see that both fine-tuning (ii), adapters (iii), and their combination (iv) give improvement on the linear readout baseline (i). However, all of them are worse than the systematically optimised adaptation architecture of \shortName{}.

The ablation also compares the results using the top-1 adaptation architecture found by SPOS architecture search against our novel progressive approach that defers the final architecture selection to an episode-wise decision. Our deferred architecture selection improves on fixing the top-1 architecture from meta-train, demonstrating the value of per-dataset/episode architecture selection (see also Sec~\ref{sec:further}).

\subsection{Further analysis}\label{sec:further}
The ablation study shows quantitatively the benefit of adaptation architecture search over common fixed adaptation strategies. In this Section, we aim to analyse: What kind of adaptation architecture is discovered by our NAS strategy, and how it is discovered? 

\noindent\textbf{Discovered Architectures}\quad
We first summarise results of the entire search space in terms of which layers are preferential to fine-tune or not, and which layers are preferential to insert adapters or not in Figure~\ref{subfig:supernet-summary}. The blocks indicate layers (columns) and adapters/fine-tuning (rows), with the color indicating whether that architectural decision was positively (green) or negatively (red) correlated with validation performance. We can see that the result is complex, without a simple pattern, as assumed by existing work~\cite{cvpr22_pmf,cvpr22_tsa,tmlr22_ett}. That said, our NAS does discover some interpretable trends. For example, adapters should be included at early/late ResNet-18 layers and not at layers 5-9. 

We next show the top three performing paths subject to diversity constraint in Figure~\ref{subfig:supernet-top3-diverse}. We see that these follow the strong trends in the search space from Figure~\ref{subfig:supernet-summary}. For example, they always adapt ($\alpha$) block 14 and never adapt block 9. However, otherwise they do include diverse decisions (such as whether to fine-tune ($\phi'$) block 15) which was not strongly indicated in Figure~\ref{subfig:supernet-summary}. 

\begin{table}[t]
  \centering
  \resizebox{1.\linewidth}{!}{
    \begin{tabular}{l|ccc}
    & \includegraphics[width=.3\linewidth]{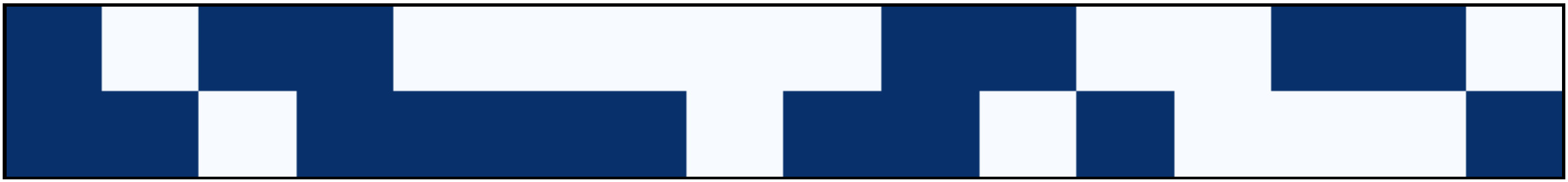} & \includegraphics[width=.3\linewidth]{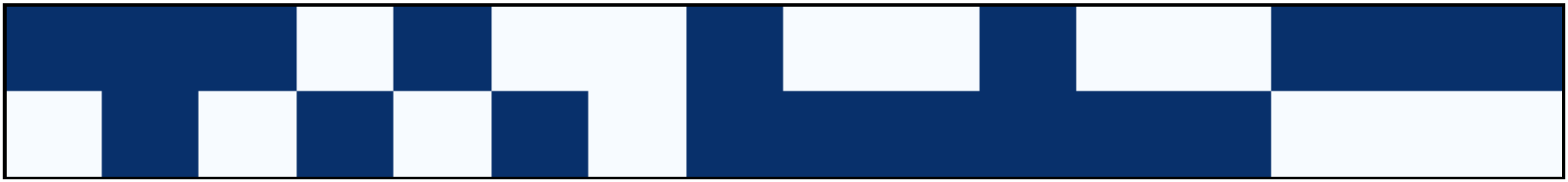} & \includegraphics[width=.3\linewidth]{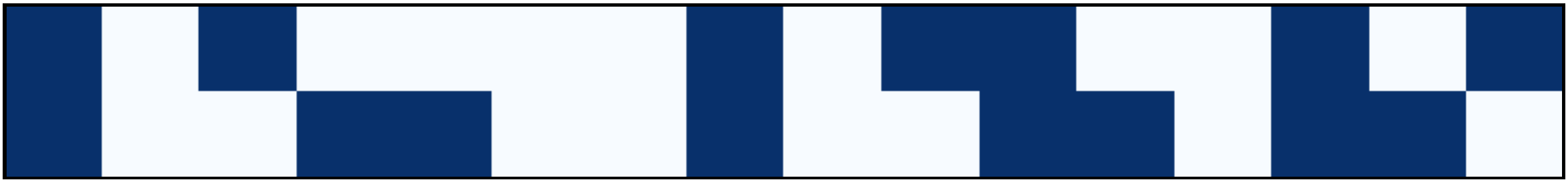} \\
    \toprule
    CIFAR-10 & \cellcolor{myGreen!26} 82.0 & \cellcolor{myGreen!7} 81.2 & \cellcolor{myGreen!66} \textbf{83.3} \\
    CIFAR-100 & \cellcolor{myGreen!85} \textbf{75.9} & \cellcolor{myGreen!11} 75.0 & \cellcolor{myGreen!3} 75.1 \\
    MNIST & \cellcolor{myGreen!90} \textbf{95.5} & \cellcolor{myGreen!10} 94.4 & \cellcolor{myGreen!0} 95.1 \\
    MSCOCO & \cellcolor{myGreen!68} \textbf{58.1} & \cellcolor{myGreen!16} 57.8 & \cellcolor{myGreen!15} 56.4 \\
    Tr. Signs & \cellcolor{myGreen!18} 81.7 & \cellcolor{myGreen!79} \textbf{82.2} & \cellcolor{myGreen!2} 81.8 \\
    \bottomrule
    \end{tabular}
  }
  \caption{How the diverse selection of architectures from Fig.~\ref{subfig:supernet-top3-diverse} perform per unseen downstream domain in Meta-Dataset. Shading indicates episode-wise architecture selection frequency, numbers indicate accuracy using the corresponding architecture. The best dataset-wise architecture (bold) is most often selected (shading).}
  \label{tab:per-domain-analysis}
  \vspace{-1em}
\end{table}

Finally, we analyse how our small set of $N=3$ candidate architectures in Figure~\ref{subfig:supernet-top3-diverse} as used during meta-test. Recall that this small set allows us to perform an efficient minimal episode-wise NAS, including for novel datasets unseen during training. The results in Table~\ref{tab:per-domain-analysis} show how often each architecture is selected by held out datasets during meta-test (shading), and what is the per-dataset performance using only that architecture. It shows how our approach successfully learns to select the most suitable architecture on a per-dataset basis, even for unseen datasets. This unique capability goes beyond prior work \cite{cvpr22_pmf,cvpr22_tsa,tmlr22_ett} where all domains must rely on the same adaptation strategy despite their diverse adaptation needs.

\noindent\textbf{Path Search Process}\quad
In addition, we illustrate the path search process in Figure~\ref{fig:search-qualitative}. This figure shows a 2D t-SNE projection of our $2K$-dimensional architecture search space, where the dots are candidate architectures of the evolutionary search process at different iterations. The dots are colored according to their validation accuracy. From the results we can see that: The initial set of candidates is broadly dispersed and generally low performing (left), and gradually converge toward a tighter cluster of high performing candidates (right). The top 3 performing paths subject to a diversity constraint (also illustrated in Fig.~\ref{subfig:supernet-top3-diverse}) are annotated in purple outline.

\noindent\textbf{Discussion}\quad
As analysed in Section~\ref{sec:ablation}, our approach can be used in either top-1 -- where each episode is a pure fine-tuning operation given the chosen architecture; or top-N architecture mode as discussed above -- where each episode performs a mini architecture selection based on the short listed produced during evolutionary search, as well as fine-tuning. We remark that while the latter imposes a slightly increased cost during testing ($N=3\times$ in practice), this is similar or less than competitors who repeat adaptation with different learning rates during testing~\cite{cvpr22_pmf} ($4\times$ cost), or exploit a backbone ensemble ($8\times$ cost) ~\cite{dvornik2020selecting,iclr21_urt}.
\section{Conclusions}
\label{sec:conclusions}

In this paper we present \shortName{}, a novel neural architecture-search based approach that discovers the optimal adaptation architecture for gradient-based few-shot learning. \shortName{} contains several recent strong heuristic adaptation architectures as special cases within its search space, and we show that by systematic architecture search they are all outperformed, leading to a new state-of-the-art on Meta-Dataset and Meta-Album. While in this paper we use a simple and coarse search space for easy and direct comparison to prior work's hand-designed adaptation strategies, in future work we will extend this framework to include a richer range of adaptation strategies, and a finer-granularity of search.

{\small
\bibliographystyle{ieee_fullname}
\bibliography{main}
}

\newpage
\appendix
\onecolumn
\section{Hyperparameter Setting}
\label{appendix:hyperparameters}

\begin{table}[t]
  \centering
  \resizebox{0.8\linewidth}{!}{
    \begin{tabular}{cl|ccc|cc}
    & & \multicolumn{3}{c|}{ResNet-18} & \multicolumn{2}{c}{ViT-S} \\
    & Hyperparameter & SDL (MD) & MDL (MD) & MDL (MA) & SDL (MD) & MDL (MD) \\
    \toprule
    & Backbone architecture & URL & URL & Supervised & DINO & DINO \\
    & Adapter architecture & TSA & TSA & TSA & ETT & ETT \\
    \midrule
    \multirow{8}{*}{\rotatebox{90} {TRAIN}} & Number of episodes & 50000 & 80000 & 20000 & 80000 & 160000 \\
    & Number of epochs & 1 & 1 & 1 & 1 & 1 \\
    & Optimizer & adadelta & adadelta & adadelta & adamw & adamw \\
    & Learning rate & 0.05 & 0.05 & 0.05 & 0.00007 & 0.00007 \\
    & Learning rate schedule & - & - & - & cosine & cosine \\
    & Learning rate warmup & - & - & - & linear & linear \\
    & Weight decay & 0.0001 & 0.0001 & 0.0001 & 0.01 & 0.01 \\
    & Weight decay schedule & - & - & - & cosine & cosine \\
    \midrule
    \multirow{11}{*}{\rotatebox{90} {SEARCH}} & Number of episodes & 100 & 100 & 100 & 100 & 100 \\
    & Number of epochs & 20 & 20 & 20 & 40 & 40 \\
    & Optimizer & adadelta & adadelta & adadelta & adamw & adamw \\
    & Learning rate & 0.1 & 0.1 & 0.1 & 0.000003 & 0.000003 \\
    & Weight decay & 0.0001 & 0.0001 & 0.0001 & 0.1 & 0.1 \\
    & Initial population size & 64 & 64 & 64 & 64 & 64 \\
    & Top-K crossover & 8 & 8 & 8 & 8 & 8 \\
    & Mutation chance & 5\% & 5\% & 5\% & 5\% & 5\% \\
    & Top-N paths & 3 & 3 & 3 & 3 & 3 \\
    & Diversity threshold & 0.4 & 0.4 & 0.4 & 0.2 & 0.2 \\
    \midrule
    \multirow{6}{*}{\rotatebox{90} {TEST}} & Number of episodes & 600 & 600 & 1800 & 600 & 600 \\
    & Number of epochs & 40 & 40 & 40 & 40 & 40 \\
    & Optimizer & adadelta & adadelta & adadelta & adamw & adamw \\
    & Learning rate & 0.1 & 0.1 & 0.1 & 0.000003 & 0.000003 \\
    & Weight decay & 0.0001 & 0.0001 & 0.0001 & 0.1 & 0.1 \\
    & Regulariser strength & 0.04 & 0.04 & 0.04 & - & - \\
    \bottomrule
    \end{tabular}
  }
  \caption{Hyperparameter setting for all experiments presented in Section~\ref{sec:experiments} of the main paper. The notation is as follows: SDL=Single domain learning, MDL=Multi-domain learning, MD=Meta-Dataset, MA=Meta-Album, TRAIN=Supernet training phase, SEARCH=Evolutionary search phase, TEST=Meta-test phase.}
  \label{tab:hyperparameter-setting}
\end{table}

Table~\ref{tab:hyperparameter-setting} reports the hyperparameters used for all of our experiments. Note the following clarifications:
\begin{itemize}
    \item ``Number of epochs'' refers to multiple forward passes of the same episode, while ``Number of episodes'' refers to the number of episodes sampled in total.
    \item The batch size is not mentioned, because we only conduct episodic learning, where we do not split the episode into batches, i.e., we feed the entire support and query set into our neural network architectures.
    \item Learning rate warmup, where applicable, occurs for the first 10\% of the episodes.
\end{itemize}
We further specify something important: While our strongest competitors~\cite{cvpr22_tsa,tmlr22_ett} tune their learning rates for meta-testing (e.g., TSA uses LR=0.1 for seen domains and LR=1.0 for unseen, and ETT uses a different learning rate per downstream Meta-Dataset domain), we treat meta-testing episodes as completely unknown, and use the same hyperparameters we used on the validation set during search.

\section{Detailed Ablation Study}
\label{appendix:ablation}

\begin{table}[t]
  \centering
  \resizebox{1.\linewidth}{!}{
    \begin{tabular}{cl|ccccccccccccc|c}
    \toprule
    & Method & Aircrafts & Birds & DTD & Fungi & ImageNet & Omniglot & QuickDraw & Flowers & CIFAR-10 & CIFAR-100 & MNIST & MSCOCO & Tr. Signs & Average \\
    \toprule
    \multirow{6}{*}{\rotatebox{90} {ResNet-18}} & $\phi\phantom{'},-$ & 64.5 & 69.6 & 71.1 & 41.2 & 56.4 & 74.8 & 64.2 & 84.6 & 75.0 & 63.9 & 82.1 & 55.9 & 77.7 & 67.8 \\
    & $\phi\phantom{'},\alpha$ & 69.6 & 67.7 & 75.0 & 42.5 & 59.5 & 71.3 & 64.9 & 88.8 & 77.4 & 70.0 & 90.2 & 58.4 & 80.1 & 70.4 \\
    & $\phi',-$ & 69.9 & 74.7 & 73.3 & 39.5 & 57.3 & 71.9 & 65.4 & 89.0 & 76.5 & 66.3 & 93.6 & 54.4 & 81.4 & 70.2 \\
    & $\phi', \alpha$ & 67.6 & 69.1 & 77.0 & 39.3 & 59.7 & 77.8 & 66.1 & 87.4 & 81.7 & 69.5 & 91.9 & 55.1 & 78.7 & 70.8 \\
    & \shortName{}-1 & 73.2 & 76.5 & 81.6 & 42.1 & 61.3 & 80.2 & 66.9 & 90.0 & 82.9 & 68.8 & 94.0 & 58.4 & 80.6 & 73.6 \\
    & \shortName{}-K & 74.9 & 76.5 & 81.6 & 50.5 & 62.7 & 80.2 & 67.2 & 94.5 & 83.0 & 71.5 & 94.0 & 59.7 & 81.9 & 75.2 \\
    \midrule
    \multirow{6}{*}{\rotatebox{90} {ViT-S}} & $\phi\phantom{'},-$ & 73.4 & 73.6 & 81.6 & 56.3 & 60.3 & 69.4 & 70.8 & 90.4 & 70.4 & 61.5 & 83.8 & 60.5 & 81.7 & 71.8 \\
    & $\phi\phantom{'},\alpha$ & 76.9 & 83.2 & 86.7 & 59.3 & 63.7 & 75.8 & 65.1 & 89.5 & 70.7 & 67.4 & 81.1 & 54.8 & 82.9 & 73.8 \\
    & $\phi',-$ & 76.8 & 80.9 & 85.8 & 61.4 & 65.9 & 73.2 & 68.5 & 91.0 & 69.9 & 66.1 & 82.5 & 57.6 & 78.8 & 74.0 \\
    & $\phi', \alpha$ & 77.0 & 83.4 & 82.4 & 58.6 & 66.7 & 73.1 & 65.0 & 95.9 & 76.7 & 66.1 & 87.7 & 58.7 & 82.9 & 74.4 \\
    & \shortName{}-1 & 83.0 & 85.5 & 87.3 & 62.2 & 68.8 & 81.9 & 72.9 & 95.3 & 79.4 & 72.6 & 95.2 & 62.6 & 87.5 & 78.7  \\
    & \shortName{}-N & 83.0 & 85.5 & 87.6 & 62.2 & 71.0 & 81.9 & 74.5 & 96.0 & 79.4 & 72.6 & 95.2 & 62.6 & 87.9 & 79.2 \\
    \bottomrule
    \end{tabular}
  }
  \caption{Ablation study on Meta-Dataset comparing four special cases of the search space: (i) $\phi,-$: No adaptation, no fine-tuning, (ii) $\phi,\alpha$: Adapt all, (iii) $\phi',-$: Fine-tune all, (iv) $\phi', \alpha$: Adapt and fine-tune all. \shortName{}-\{1,N\} refer to conventional and deferred episode-wise NAS respectively. Single domain setting: Only ImageNet is seen during training and search. Reporting mean accuracy over 600 episodes.}
  \label{tab:ablation-single-domain}
\end{table}

\begin{table}[t]
  \centering
  \resizebox{1.\linewidth}{!}{
    \begin{tabular}{cl|ccccccccccccc|c}
    \toprule
    & Method & Aircrafts & Birds & DTD & Fungi & ImageNet & Omniglot & QuickDraw & Flowers & CIFAR-10 & CIFAR-100 & MNIST & MSCOCO & Tr. Signs & Average \\
    \toprule
    \multirow{6}{*}{\rotatebox{90} {ResNet-18}} & $\phi\phantom{'},-$ & 64.5 & 69.6 & 71.1 & 41.2 & 56.4 & 74.8 & 64.2 & 84.6 & 75.0 & 63.9 & 82.1 & 55.9 & 77.7 & 67.8 \\
    & $\phi\phantom{'},\alpha$ & 89.3 & 78.3 & 76.1 & 62.7 & 57.2 & 93.8 & 76.0 & 90.8 & 77.8 & 66.1 & 90.5 & 56.9 & 79.5 & 76.5 \\
    & $\phi',-$ & 90.2 & 76.7 & 70.6 & 63.1 & 57.8 & 88.2 & 79.3 & 88.9 & 78.2 & 68.2 & 96.1 & 51.7 & 82.9 & 76.3 \\
    & $\phi', \alpha$ & 86.1 & 78.9 & 77.2 & 60.5 & 57.6 & 94.1 & 79.5 & 86.5 & 81.0 & 67.2 & 96.1 & 52.6 & 81.8 & 76.9 \\
    & \shortName{}-1 & 90.1 & 82.1 & 79.9 & 67.9 & 61.4 & 94.3 & 82.6 & 92.2 & 82.4 & 73.8 & 95.4 & 58.1 & 81.0 & 80.1 \\
    & \shortName{}-K & 90.1 & 83.8 & 82.3 & 68.4 & 61.4 & 94.3 & 82.6 & 92.2 & 83.0 & 75.1 & 95.4 & 58.8 & 81.9 & 80.7 \\
    \midrule
    \multirow{6}{*}{\rotatebox{90} {ViT-S}} & $\phi\phantom{'},-$ & 73.4 & 73.6 & 81.6 & 56.3 & 60.3 & 69.4 & 70.8 & 90.4 & 70.4 & 61.5 & 83.8 & 60.5 & 81.7 & 71.8 \\
    & $\phi\phantom{'},\alpha$ & 85.7 & 84.3 & 81.8 & 68.7 & 70.4 & 89.1 & 77.0 & 90.2 & 73.5 & 61.4 & 82.6 & 53.7 & 72.4 & 77.3 \\
    & $\phi',-$ & 83.0 & 84.5 & 81.1 & 70.9 & 72.4 & 88.6 & 74.6 & 90.4 & 75.1 & 63.5 & 87.0 & 54.0 & 75.5 & 77.5 \\
    & $\phi', \alpha$ & 82.5 & 85.9 & 82.7 & 68.9 & 73.7 & 90.4 & 77.1 & 94.0 & 73.4 & 66.2 & 85.9 & 55.9 & 77.4 & 78.9 \\
    & \shortName{}-1 & 89.1 & 90.3 & 86.3 & 75.1 & 74.6 & 92.0 & 80.6 & 93.5 & 75.9 & 70.8 & 91.3 & 62.7 & 87.2 & 83.1 \\
    & \shortName{}-N & 89.1 & 92.5 & 86.3 & 75.1 & 74.6 & 92.0 & 80.6 & 93.5 & 75.9 & 70.8 & 91.3 & 62.8 & 87.2 & 83.4 \\
    \bottomrule
    \end{tabular}
  }
  \caption{Ablation study on Meta-Dataset comparing four special cases of the search space: (i) $\phi,-$: No adaptation, no fine-tuning, (ii) $\phi,\alpha$: Adapt all, (iii) $\phi',-$: Fine-tune all, (iv) $\phi', \alpha$: Adapt and fine-tune all. \shortName{}-\{1,N\} refer to conventional and deferred episode-wise NAS respectively. Multi-domain setting: The first 8 datasets are seen during training and search. Reporting mean accuracy over 600 episodes.}
  \label{tab:ablation-multi-domain}
\end{table}

Tables~\ref{tab:ablation-single-domain} and~\ref{tab:ablation-multi-domain} provide the exact scores per Meta-Dataset domain that are summarised in Table~\ref{tab:ablation-short} of the main paper, for single domain and multi-domain FSL respectively.

\section{Source code}
\label{appendix:source}

The source code is available at: \url{https://github.com/peustr/nfts-public}.

\end{document}